\documentclass[journal]{IEEEtran}

\usepackage{algorithm}
\usepackage{algorithmicx}
\usepackage[noend]{algpseudocode} 

\usepackage{acro}
\usepackage{graphicx}
\usepackage{float}
\usepackage{xcolor}
\usepackage{amsmath}
\usepackage{amssymb}
\usepackage{hyperref}
\hypersetup{
    colorlinks=true,
    linkcolor=blue,
    filecolor=magenta,      
    urlcolor=red,
}

\newcommand{\Fig}{Fig.}

\newcommand{\Tab}{Tab.}

\DeclareAcronym{WearMask}{
short=WearMask,
long=Web-based efficient AI recognition of masks,
}

\DeclareAcronym{YOLO}{
short=YOLO,
long=You only look once,
}

\DeclareAcronym{WASM}{
short=WASM,
long=WebAssembly,
}

\DeclareAcronym{FPS}{
short=FPS,
long=Frames Per Second,
}

\DeclareAcronym{MAFA}{
short=MAFA,
long=MAsked FAces,
}

\DeclareAcronym{RMFD}{
short=RMFD,
long=Real-World Masked Face Dataset,
}

\DeclareAcronym{SMFD}{
short=SMFD,
long=Simulated Masked Face Dataset,
}

\DeclareAcronym{AP}{
short=AP,
long=Average Precision,
}

\DeclareAcronym{mAP@0.5}{
short=mAP@0.5,
long=mean of Average Precision when IoU is 0.5,
}

\DeclareAcronym{IoU}{
short=IoU,
long=Intersection Over Union,
}

\begin{document}
%
\title{WearMask: Fast In-browser Face Mask Detection with Serverless Edge Computing for COVID-19}
%
%
%

\author{Zekun~Wang,
        Pengwei~Wang,
        Peter~C.~Louis,
        Lee~E.~Wheless,
        and~Yuankai~Huo,~\IEEEmembership{Member,~IEEE}
\thanks{Z. Wang, Y. Huo were with the Data Science Institute, Vanderbilt University, Nashville,
TN, 37215, USA}
\thanks{P. Wang is the corresponding author, was with the Department of Information Science and Engineering, Shandong University, Qingdao, China,  e-mail: wangpw@sdu.edu.cn}
\thanks{P. Louis was with the Department of Biomedical Informatics, Vanderbilt University Medical Center, Nashville,TN, 37215, USA}
\thanks{L. Wheless was with the Department of Dermatology, Vanderbilt University Medical Center, Nashville,TN, 37215, USA}
\thanks{Y. Huo was with the Department of Computer Science, Vanderbilt University, Nashville,
TN, 37215, USA}
}

\markboth{Manuscript pre-print, December~2020}%
{Shell \MakeLowercase{\textit{et al.}}: Bare Demo of IEEEtran.cls for IEEE Journals}

\maketitle

\begin{abstract}
The COVID-19 epidemic has been a significant healthcare challenge in the United States. According to the Centers for Disease Control and Prevention (CDC), COVID-19 infection is transmitted predominately by respiratory droplets generated when people breathe, talk, cough, or sneeze. Wearing a mask is the primary, effective, and convenient method of blocking 80\% of all respiratory infections. Therefore, many face mask detection and monitoring systems have been developed to provide effective supervision for hospitals, airports, publication transportation, sports venues, and retail locations. However, the current commercial face mask detection systems are typically bundled with specific software or hardware, impeding public accessibility. In this paper, we propose an in-browser serverless edge-computing based face mask detection solution, called Web-based efficient AI recognition of masks (WearMask), which can be deployed on any common devices (e.g., cell phones, tablets, computers) that have internet connections using web browsers, without installing any software. The serverless edge-computing design minimizes the extra hardware costs (e.g., specific devices or cloud computing servers). The contribution of the proposed method is to provide a holistic edge-computing framework of integrating (1) deep learning models (YOLO), (2) high-performance neural network inference computing framework (NCNN), and (3) a stack-based virtual machine (WebAssembly). For end-users, our web-based solution has advantages of (1) serverless edge-computing design with minimal device limitation and privacy risk, (2) installation free deployment, (3) low computing requirements, and (4) high detection speed. Our WearMask application has been launched with public access at \url{facemask-detection.com}.
\end{abstract}


%
\IEEEpeerreviewmaketitle

\begin{figure*}[t]
\begin{center}
\includegraphics[width=1\textwidth]{{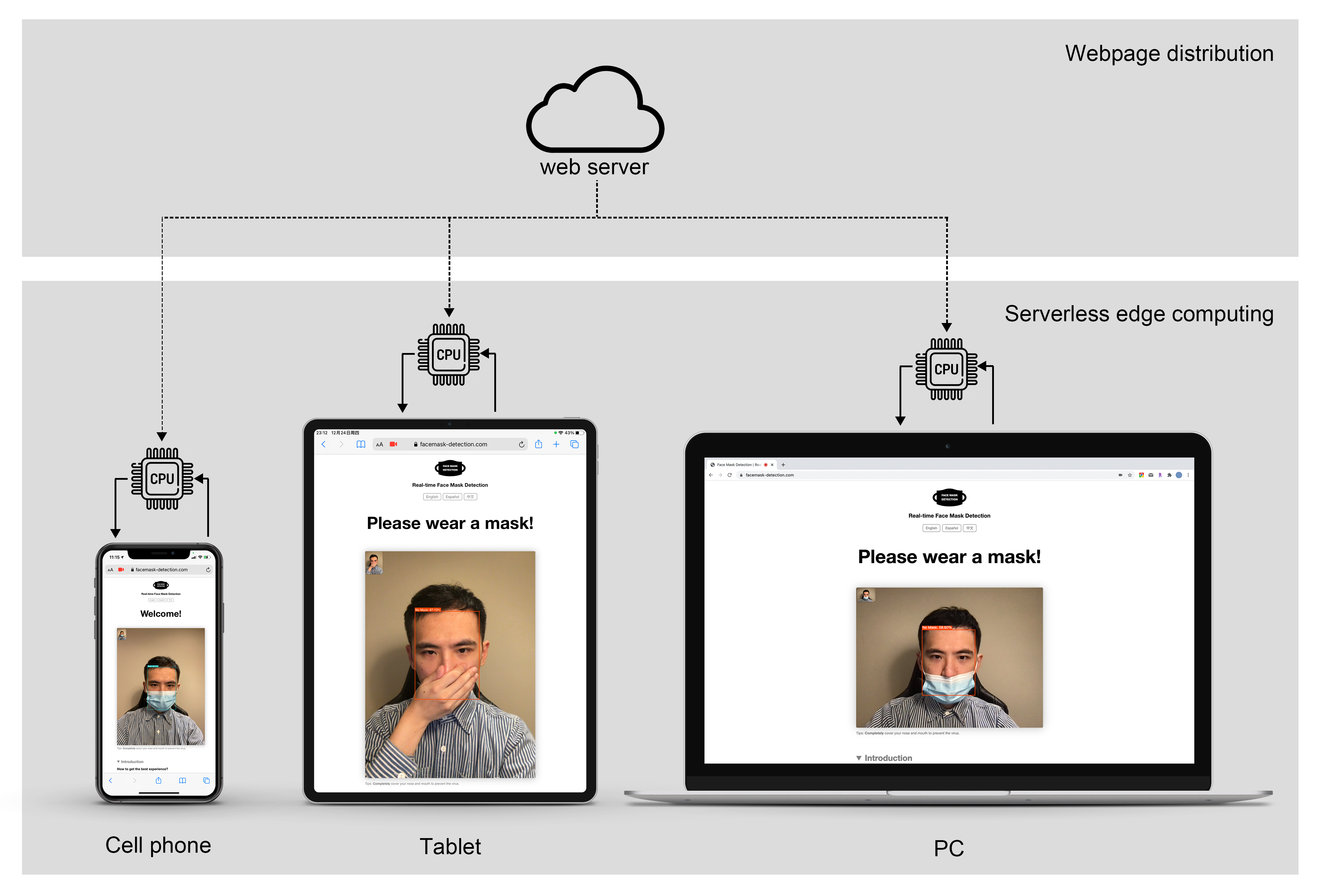}}
\end{center}
\caption{This figure shows the connection between the web server and edge devices.} 
\label{Fig.1} 
\end{figure*}

\begin{table*}
\caption{Solution comparison}
\centering
\begin{tabular}{cccccc}
\hline
Solution&Amazon Recognition&Mask detector&MaskDetection&Mask detection&WearMask\\
 ~&PPE detection&(Android APP)&(Kogniz.ai)&machine&(Our solution)\\
 
\hline
  No specific hardware&$\checkmark$&$\checkmark$&$\times$&$\times$&$\checkmark$\\
  No installation&$\times$&$\times$&$\checkmark$&$\checkmark$&$\checkmark$\\
  Cross platform&$\checkmark$&$\times$&$\checkmark$&$\times$&$\checkmark$\\
  Local computing&$\times$&$\checkmark$&$\times$&$\checkmark$&$\checkmark$\\
  
\hline
\end{tabular}
\label{table1} 
\end{table*}

\section{Introduction}
\IEEEPARstart{S}{ince} November 2019, the COVID-19 epidemic had been a major social and healthcare issue in the United States. Only during the Thanksgiving week in 2020, there were 1,147,489 new confirmed cases and 10,279 new deaths from COVID-19~\cite{1p3a2020data}. It is necessary to wear masks in public places~\cite{chu2020physical}. Even with the successful development of many vaccines, wearing a mask is still one of the most effective and affordable ways to block 80\% of all respiratory infections and cut off the route of transmission~\cite{centers2020scientific}. Even many states have enforced people to wear masks in public places, there are still a considerable number of people who forget or refuse to wear masks, or wear masks improperly. Such facts would increase the infection rate and eventually bring a heavier load of the public health care system. Therefore, many face mask monitoring systems have been developed to provide effective supervision for hospitals, airports, publication transportation systems, sports venues, and retail locations. 

However, the current commercial face mask detection systems are typically bundled with specific software or hardware, impeding public accessibility. Herein, it would be appealing to design a light-weight device agnostic solution to enable fast and convenient face mask detection deployment. In this paper, we propose a serverless edge-computing based in-browser face mask detection solution, called \ac{WearMask}, which can be deployed on any common devices (e.g., cell phones, tablets, computers) that have an internet connection and a web browser, without installing any software. 

Serverless edge-computing is a recent infrastructural evolution of edge-computing, in which computing resources are directly used by end-users. The features of existing computing strategies are listed in {\color{red} \Tab~\ref{table1}}. To maximize the flexibility and convenience of the deployment for small business, the serverless edge-computing design is introduced in our \ac{WearMask} framework. As opposed to canonical edge computing, serverless edge-computing does not require extra hardware between a web-server and end-users. Web browsers (e.g., Chrome and Firefox) are used as the interfaces since they are the most widely accessible interface for users to access the internet, which is device and operating system (OS) agnostic. On the other hand, most internet users are familiar with web browsers, which introduce almost no extra learning burdens for deploying our \ac{WearMask} software. To do so, we aggregate a holistic solution by combining serverless edge-computing and deep learning based object detection, without the requirement of having advanced GPU.

The technical contribution of the proposed method is to provide a holistic serverless edge-computing framework with (1) deep learning models (YOLO~\cite{redmon2016you}), (2) high-performance neural network inference computing framework (NCNN~\cite{nihui2020ncnn}), and (3) a format running on the stack-based virtual machine (WebAssembly~\cite{rossberg2018bringing}). For end-users, the advantages of the proposed web-based solution are (1) minimal device limitation, (2) installation free design, (3) low computing requirements, and (4) high detection speed. Our \ac{WearMask} application has been launched with public access as a website \footnote{facemask-detection.com} ({\color{red} \Fig~\ref{Fig.1}}).

Due to the light-weight device-agnostic design, the proposed method would be a cost-efficient solution for public facilities and small businesses. For example, to set up a commercial mask detection solution, averagely costs \$1,000 - \$4,000, leading to an extra burden for small businesses, especially considering the financial challenges during the pandemic. As the United States has more than 30.2 million small businesses~\cite{sba2018small} and a large number of public facilities, it is essential to find an affordable alternative.


The remainder of the paper is organized as follows. First, we review recent works in face mask detection. Then, we discuss the data curation, model selection, and the training process (Yolo-Fastest~\cite{dogqiuqiu2020yolofastest}). Next, we describe the specific deployment process of the model (NCNN, WebAssembly). Finally, we discuss this scheme's existing strengths and weaknesses and other deployment schemes' advantages and disadvantages.


\begin{figure*}
\begin{center}
\includegraphics[width=1\textwidth]{{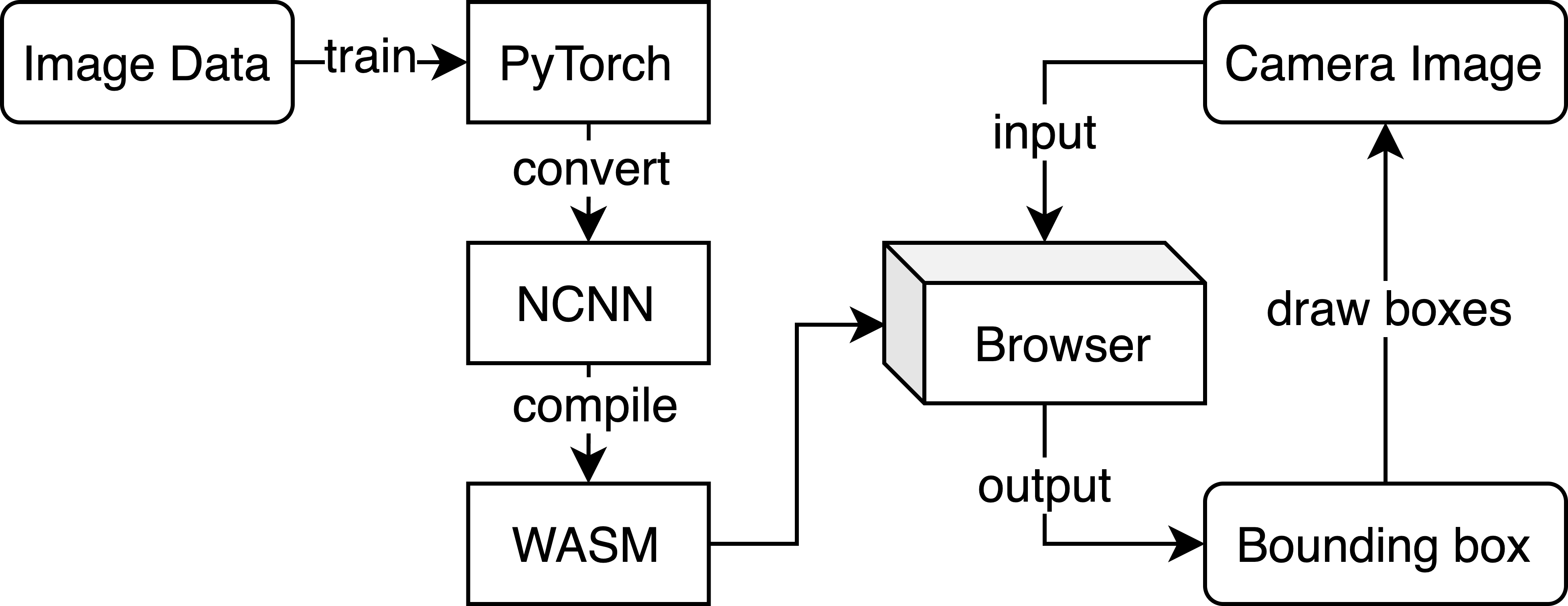}}
\end{center}
\caption{This figure shows the overall workflow of training and deploying of the proposed WearMask framework.} 
\label{Fig.2} 
\end{figure*}

\section{Related Works}
\subsection{Face Mask Detection}
Historically, most of the papers focused on performing face recognition while wearing a mask or with other obstructions, while few, if any, previous studies have been conducted to determine if a subject is wearing a mask. Since the COVID-19 pandemic, however, more efforts have been devoted to face mask detection due to the emergence of the need for reducing COVID-19 transmission.

To detect occluded faces, Shiming Ge et al. established the \ac{MAFA} dataset, which contains 30,811 images and established the LLE-CNNs model, obtaining an Average Precision (AP) of 76.4\%~\cite{ge2017detecting}. A. Nieto-Rodríguez and others developed an alarm system for the wearing of masks in the operating room, combining the face detector with the mask detector, and optimized for low False Positive rate and high Recall, and obtained 95\% True Positive rate~\cite{nieto2015system}. But its direction is limited to the surgical mask in the operating room. Bosheng Qin et al. established SRCNet and classified the mask-wearing situation into three categories: no facemask-wearing, incorrect facemask-wearing, and correct facemask-wearing, to classify images and obtain 98.70\% accuracy~\cite{qin2020identifying}. However, it needs to crop the face area before face mask detection and concentrate on the face. Mohamed Loey et al. used ResNet-50 and SVM to obtain 99.49\% accuracy on \ac{RMFD}~\cite{loey2020hybrid}. Later, they used YOLO-V2 based on ResNet-50 for object detection in another publication and achieved 81\% \ac{AP} on the Medical Masks Dataset (MMD) and Face Mask Dataset (FMD)~\cite{loey2020fighting}. G. Jignesh Chowdary et al. achieved 99.9\% accuracy on \ac{SMFD} with an InceptionV3 based model~\cite{chowdary2020face}. Nevertheless, since \ac{SMFD} is a small dataset generated using mask images. Results are more difficult to be generalized in complicated scenarios.

As opposed to these studies, whose goals were to develop a more precise face mask detection algorithm, our work emphasizes leveraging lightweight serverless edge-computing based in-browser deployment of face mask detection with deep learning. To the best of our knowledge, no previous publications have investigated the serverless edge-computing based in-browser solution of face mask detection by aggregating deep learning models, serverless edge-computing frameworks, and stack-based virtual machines.


\subsection{In-browser Serverless Edge Computing}
The typical way of designing in-browser deep learning model is to use TensorFlow.js~\cite{smilkov2019tensorflow} or ONNX.js~\cite{microsoft2020onnxjs}. These methods use a specialized JavaScript library to read the model file and completes the inference through calling computing operations written in JavaScript. However, JavaScript is not a typical programming language in the deep learning field, with less community support.

\subsubsection{ONNX.js}
To use ONNX.js, we first need to convert the existing PyTorch~\cite{paszke2019pytorch} model into an ONNX (Open Neural Network Exchange)~\cite{bai2019onnx} model. ONNX defines various standard operators in machine learning and deep learning as an open format to facilitate developers to use ONNX as a relay to convert models from one framework to another. Now ONNX has supported PyTorch, TensorFlow~\cite{abadi2016tensorflow}, Caffe2, NCNN, and other common deep learning frameworks. ONNX.js is a JavaScript library that can directly read the ONNX model in the JavaScript environment for inference. However, there are limitations to complete in-browser deployments using ONNX.js:
(1) It does not support INT64 format variables in the model. Since ONNX.js runs in the JavaScript environment and JavaScript does not support INT64 format variables. The ONNX model, directly exported by PyTorch, contains many variables in the INT64 format. Therefore, we need to convert the INT64 component to INT32 format in the ONNX model. Nevertheless, even after replacing all variables with INT32 format, the ONNX model still does not work correctly. Some native ONNX operators only support INT64 as input to the node, such as ConstantOfShape. The official ONNX model interpreter no longer supports the modified operator.
(2) Many operators are not supported. ONNX.js needs to read the model and call its built-in JavaScript operations based on the model content, which means that the ONNX.js library must have defined all operators before performing them. Because ONNX.js is a new niche, its supporting operators are relatively few. For example, the Resize operation is not supported. Considering the rapid emergence of new models and ideas nowadays, this is a significant limitation for model deployment.

\subsubsection{TensorFlow.js}
Compared to ONNX.js, TensorFlow.js is more widely used and supports a richer set of operators. To use TensorFlow.js, we first convert the model to a TensorFlow SavedModel format. Since there is no official way to convert a PyTorch model to TensorFlow, we adapt it to an ONNX model and then convert it into a TensorFlow SavedModel. Next, We convert the SavedModel to a particular readable web format model that can be read by TensorFlow.js in a JavaScript environment. Complex conversion chains mean lower reliability and more limitations. In its transformation, it is likely to introduce some uncommon operators in the model. For example, in our attempt, this process introduces the operator called TensorScatterUpdate, which is not supported for the particular readable web format model.

\section{Methods}
\subsection{Data Collection}

The data used to train the \ac{WearMask} model consists of two types: (1) face with masks, and (2) face without masks. The normal face data (without masks) were collected from the WIDER FACE dataset~\cite{yang2016wider}. The existing mask datasets were divided into two categories, including real faces with real masks (\ac{MAFA}, \ac{RMFD}~\cite{wang2020masked}, MMD) and real faces with generated masks (\ac{SMFD}). Considering that the generated mask images influence model generalization ability, we used real images instead. Briefly, the \ac{MAFA} is composed of face pictures with various faces with heterogeneous masks, background, and annotation information. The \ac{RMFD} only contains the faces without background. Considering that the background images without faces would help the model improve precision, the \ac{MAFA} dataset were finally selected as training data. However, the face in \ac{MAFA} only contains the regions below the eyebrows, while the WIDER FACE dataset marks the whole face's position. To reconcile the differences from the annotation protocols of two datasets, the partial annotations in WIDER FACE was extended to whole faces, following the protocol that the marked bounding box range was from below the hairline to above the lower edge of the chin, and the left and right borders do not include the ears. Moreover, we also collected some samples of incorrectly wearing masks and some edge conditions. As a result, the final training dataset contained 4065 images from \ac{MAFA}, 3894 images from WIDER FACE, and 1138 additional images from the internet. In general, a total of 9,097 images with 17,532 labeled boxes were divided into 80\% training and validation, and 20\% testing.


\subsection{Modeling}
To deploy deep learning models on edge devices without advanced GPUs, the efficiency and size of such models are essential. In general, the smaller the model is, the less computational resource is needed on edge devices. The YOLO-Fastest model is employed as the detection method in the \ac{WearMask} solution, as a lightweight version of \ac{YOLO}~\cite{redmon2016you} object detection model. \ac{YOLO} is one of the most widely used fast object detection algorithms. As opposed to the traditional anchor-based detection algorithms (e.g., Faster-RCNN~\cite{ren2016faster}), \ac{YOLO} divides the image into predefined grids to match the target objects. The grid definition alleviates repetitive anchor computation, thus improves the computational efficiency dramatically. To further improve the computational speed, YOLO-Fastest was proposed as an open-source object detection model. In the YOLO-Fastest implementation, the EfficientNet-lite~\cite{tan2019efficientnet} is employed as the encoder. The total model size is only 1.3 MB (MegaByte), compatible with low computing power scenarios such as edge devices. 

To further speed up the convergence of the training process, the COCO dataset (Microsoft Common Objects in Context, 80 categories)~\cite{tan2019efficientnet} was employed to pretrain the detection backbone. The model was then further trained by the face mask datasets as a transfer learning practice~\cite{pan2009survey}, by adjusting the network to suit new object definitions.

\subsection{Serverless Edge-Computing}

Moreover, as opposed to current cloud-computing based implementations, we implemented the \ac{WearMask} as a serverless edge-computing framework. For face mask detection, the cloud computing solution is limited by the availability of enough internet bandwidth of real-time camera video streams, with high costs for cloud services. By contrast, our serverless edge-computing design minimizes the hardware costs by utilizing users’ existing devices. Beyond the lower hardware costs, the edge-computing design has another critical advantage of low privacy risk. In \ac{WearMask} framework, the video data are processed locally without uploading processes (e.g., upload to cloud servers), which is essential as a healthcare-related method.




\subsection{In-browser Deployment}
We further implement the edge-computing strategy as an in-browser deployment by aggregating NCNN and \ac{WASM} architectures ({\color{red} \Fig~\ref{Fig.2}}). NCNN is an open-source optimized inference framework for mobile platforms developed by Tencent. It is implemented in pure C++ without third-party library dependencies. Herein, the size of the compiled model will be minimized, with decent computing efficiency on edge devices, especially on devices with ARM architecture chips. It supports custom layers, which means it works when there are custom operators in the model. Moreover, it has provided tools to convert from various frameworks to NCNN. \ac{WASM} is a low-level language that run in the browser. It is in binary form, which is faster than JavaScript programs. Moreover, prevalent web browsers, such as Chrome, Edge, Firefox, and WebKit (Safari), have already supported \ac{WASM} with minimal generalizability barriers. 

Using the NCNN library, we first converted the PyTorch model into an NCNN model with a model size of 581 KB. Then we created a new C++ program to streamline the detection process from image preprocessing to finally output the category, confidence, and boxes position, using the NCNN library for inference. After compiling this C++ program into \ac{WASM} format, the entire framework was executed as a function in JavaScript. To visualize the detection results, we used HTML and CSS to render the detected bounding boxes with original face images.



        
    


\begin{figure*}
\begin{center}
\includegraphics[width=1\textwidth]{{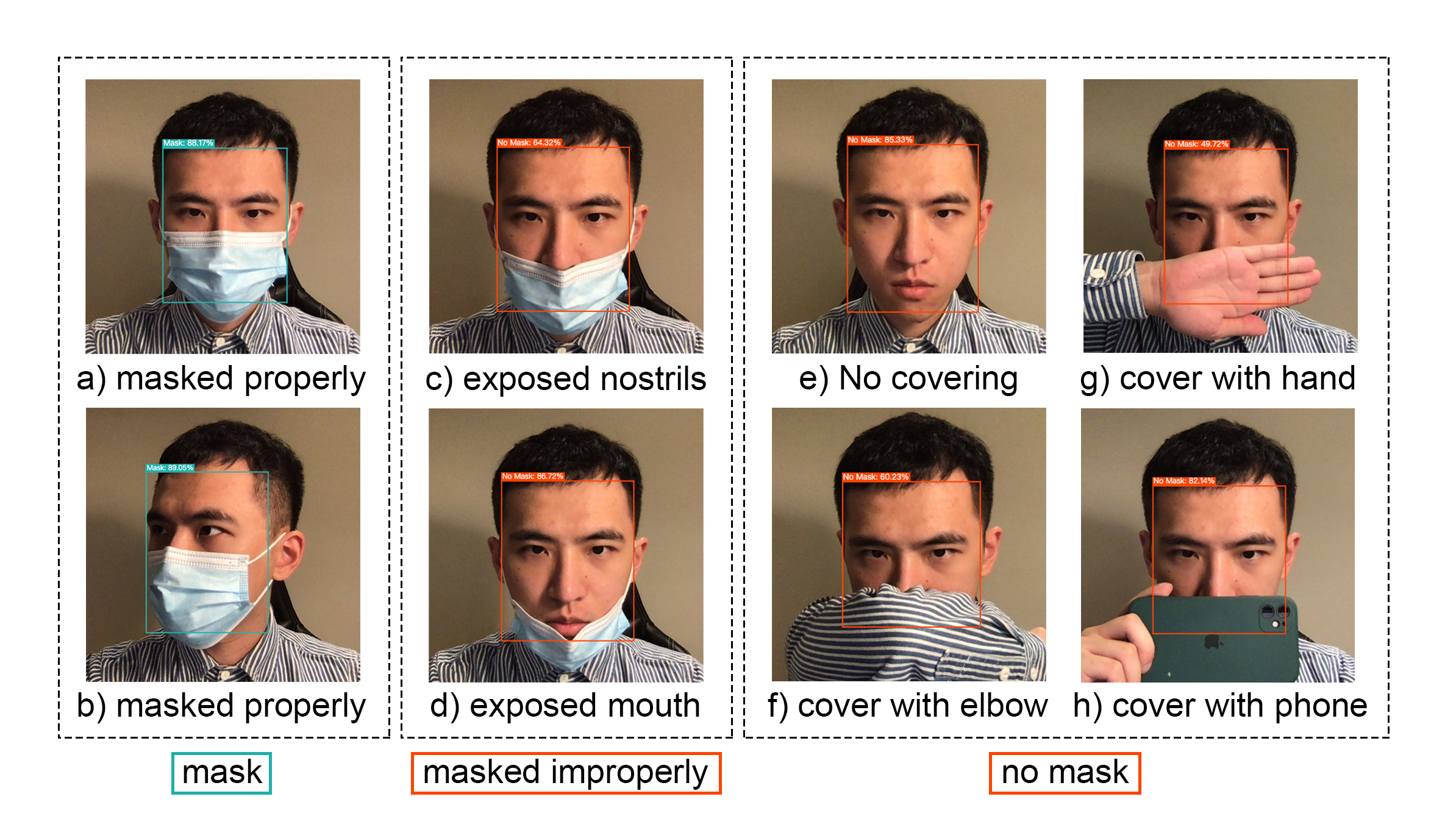}}
\end{center}
\caption{This figure shows the different detection results. a) and b) show the cases with wear masks properly, c) and d) are the examples that wear mask improperly, and e), f), g) and h) are the cases of not wearing the mask.} 
\label{Fig.3} 
\end{figure*}

\section{Experiments and Results}

\subsection{Experimental Setting}
The training was performed on Google Colab (Tesla V100-SXM2-16GB). The PyTorch Version was 1.7.0 + cu 101. The training code was modified from the public code for YOLO-V3~\cite{redmon2018yolov3} from Ultralytics~\cite{jocher2019guigarfr}.

\begin{figure}
\begin{center}
\includegraphics[width=0.45\textwidth]{{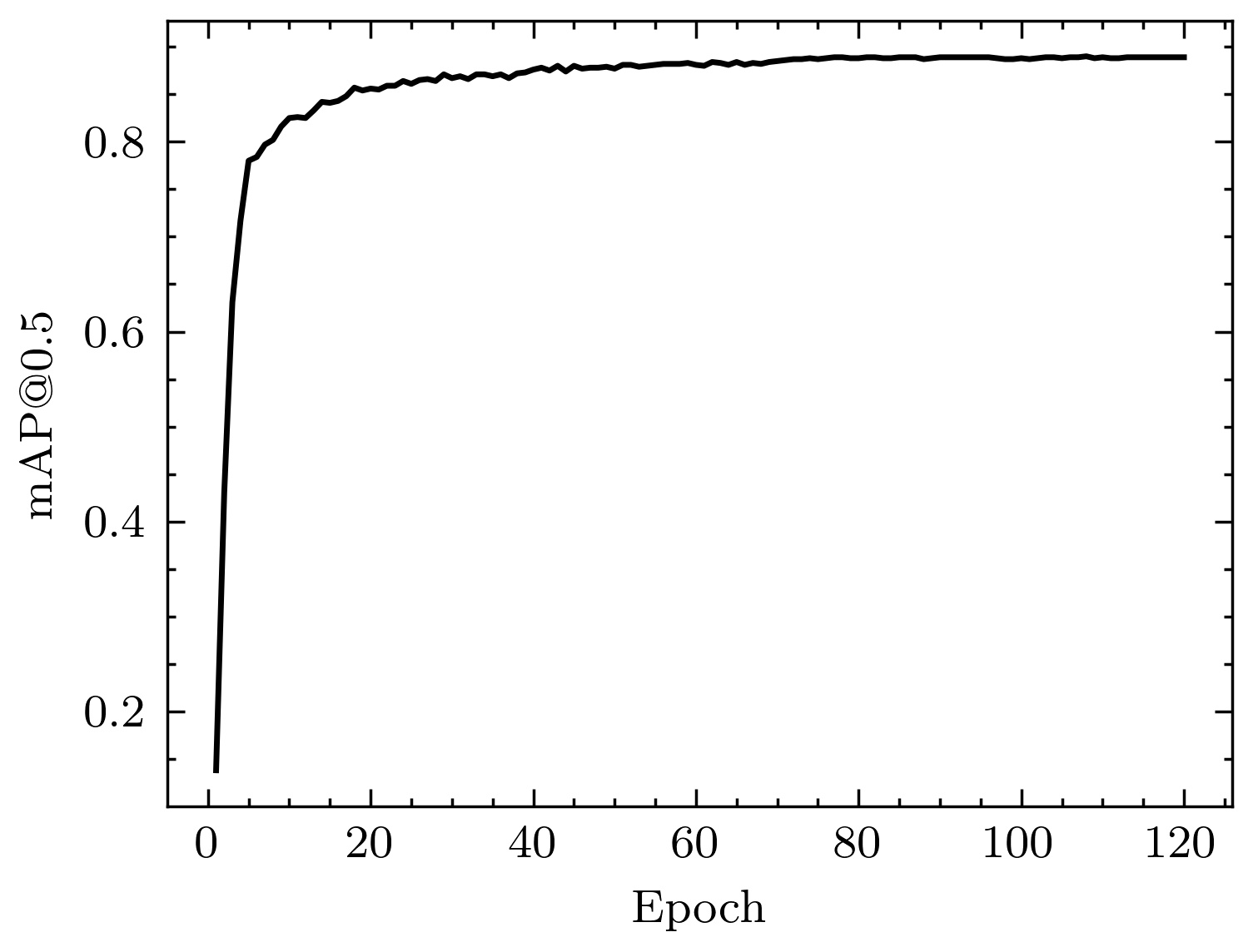}}
\end{center}
\caption{This figure shows the testing mAP @ 0.5 of different training epochs.} 
\label{Fig.5} 
\end{figure}

\begin{figure}
\begin{center}
\includegraphics[width=0.45\textwidth]{{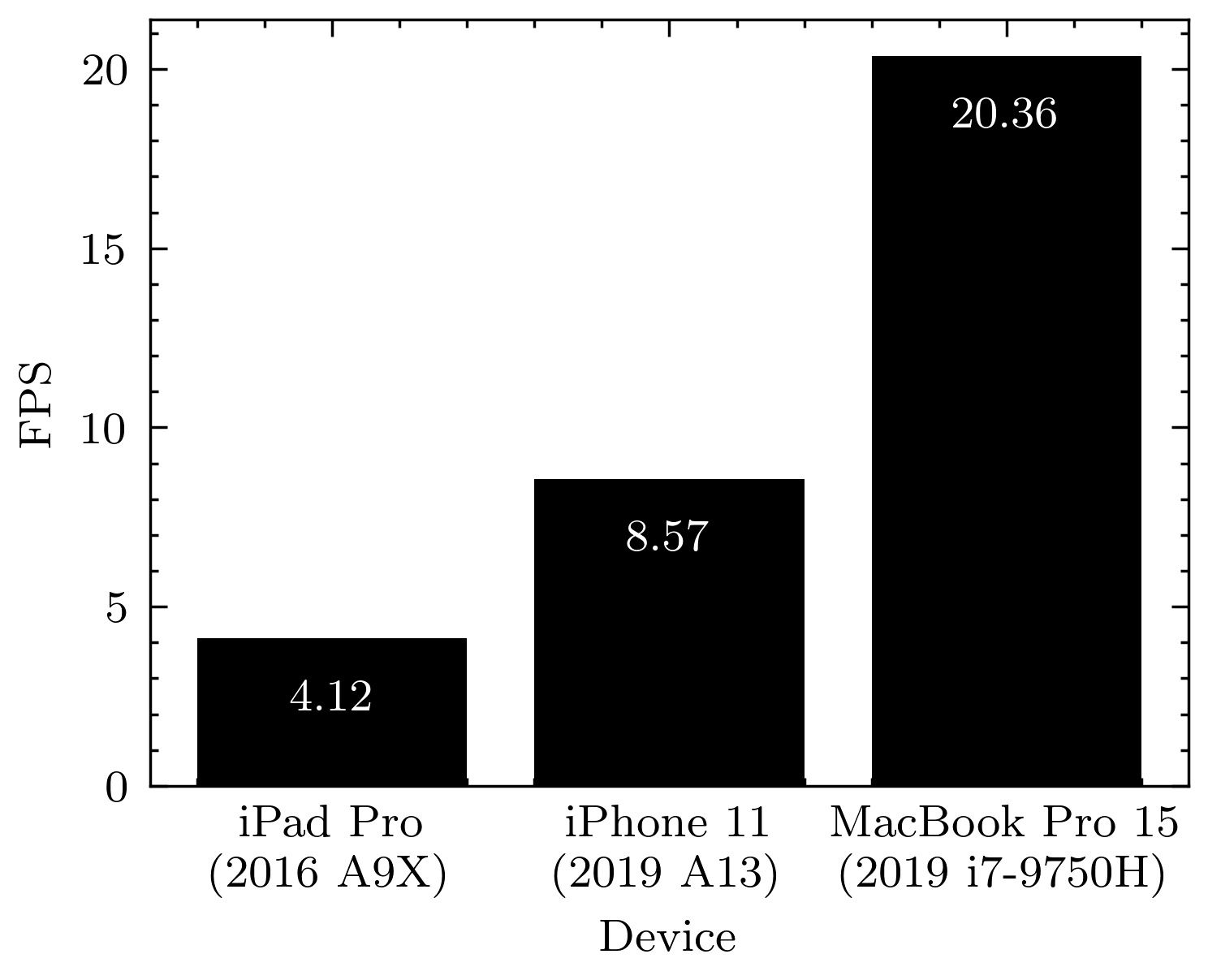}}
\end{center}
\caption{This figure shows the different Frames Per Second (FPS) using different edge devices.} 
\label{Fig.4} 
\end{figure}

\begin{figure*}
\begin{center}
\includegraphics[width=0.9\textwidth]{{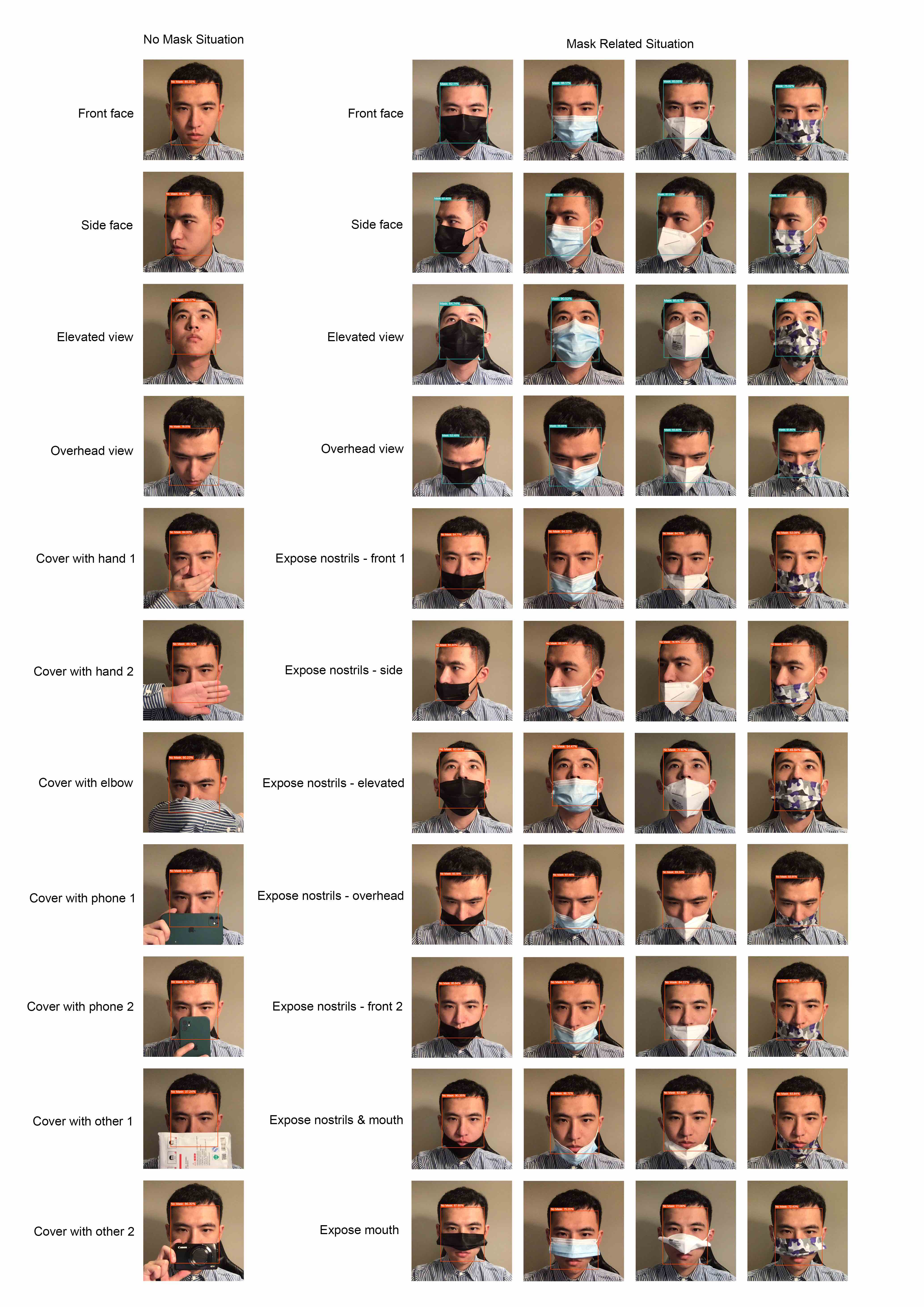}}
\end{center}
\caption{This figure shows more testing examples using the proposed WearMask tool.} 
\label{Fig.6} 
\end{figure*}

\subsection{Results}
The qualitative results of our proposed WearMask framework is presented in ({\color{red} \Fig~\ref{Fig.3}}).  It is able to correctly identify the cases when a subject is not wearing the mask properly, such as when the nostrils or mouth are exposed. It also has a higher probability of recognizing that it is not a mask even if the object uses the hands, elbows, or other things such as cell phones to cover the face. The model is able to work on multiple faces within the same frame.

From quantitative results, the trained YOLO-Fastest model achieved \ac{mAP@0.5} of 0.89 after 120 epochs with batch-size 16({\color{red} \Fig~\ref{Fig.5}}). The detection FPS on representative edge devices are presented is {\color{red} \Fig~\ref{Fig.4}}.

After converting the PyTorch model to an NCNN model with \ac{WASM}, we established a website and published it at \url{facemask-detection.com},  to demo the in-browser deployment.

\subsection{Comparison with Previous Studies}

Previous related works used various datasets with different tasks and evaluation strategies. For instance, some of the methods evaluated the performance of detection and classification using detection metrics, while other works only evaluated detection or classification, respectively. Therefore, most of the previous evaluation results cannot be directly compared with the results obtained in this paper. In this study, we compared our performance with the most similar settings among previous works, which (1) used real mask pictures instead of simulated mask pictures, (2) performed simultaneous object detection and classification. 

Among them, the LLE-CNNs model used by Shiming Ge et al. obtained an AP of 76.4\%~\cite{ge2017detecting}. Mohamed Loey et al. used YOLOV2 with ResNet-50 and obtained an AP of 81\%~\cite{loey2020fighting}. Our model achieved an AP of 89\% ({\color{red} \Fig~\ref{Fig.5}}). Although the datasets are not completely same, it shows that the performance of this model is at a comparable level with prior arts. Moreover, this study aims to build a light-weight edge computing framework for face mask detection rather than to achieve the best detection accuracy. More testing examples are presented in ({\color{red} \Fig~\ref{Fig.6}})


\section{Discussion}
Current mask detection solutions typically require specialized equipment or dedicated environments for deployments, which are often unscalable, high cost, or not flexible. To address such issues, we proposed a new edge-computing based face mask detection method, called \ac{WearMask}, with the following features:

(1) Serverless edge-computing design. The proposed framework can be deployed conveniently with minimal costs and high flexibility. The deep learning model size is minimized for edge devices. 

(2) Easy deployments. The framework is device-agnostic (e.g., across computer, laptop, cell phone, or tablet) and compatible with major operating systems (e.g., Windows, macOS, Linux, Android, and iOS). 

(3) Installation free. No installation process and environmental settings are required to use the application. Users only need to visit our web page and enable camera permissions to trigger the software. 

(4) Low privacy risk. Since the program is run locally without exchanging data, there was no need to save any contents or upload any data to the server. For privacy purposes, the model and data can be disconnected from the Internet after they are loaded. 


The proposed mask detection solution also had several limitations. Since most common devices do not support infrared detection, the program cannot detect human body temperature as professional equipment. As a reminder-only device, it cannot force a person to wear a mask if they insist on not wearing it.

The potential future improvements of this mask detection scheme will be as follows: 
(1) The existing dataset divides the no mask and the wearing masks incorrectly into one category, which causes some misunderstanding in detection. Fine-grain classification would be helpful with more training data.
(2) For a subject that does not wear the mask properly, the current scheme can recognize this case. However, it cannot remind the subject of the specific incorrect location, such as revealing the nostril or mouth. In the future, we can add an attention map~\cite{schlemper2018attention} to specify the location that the mask is not worn properly.
(3) Due to the system limitation in iOS, only Safari supports WebAssembly-related functions, and it does not support parallel computing features such as SIMD (Single instruction, multiple data). Therefore, when running on iOS, the \ac{FPS} is significantly lower than the performance of the same CPU level on other system devices. From our experiments, when SIMD-related features were enabled, the \ac{FPS} was twice faster. 

\section{Conclusion}
In this paper, we proposed a system agnostic no-installation face mask detection solution to remind people who are not wearing a mask or wearing a mask improperly. As a serverless edge-computing design, it can be run locally on various edge devices, with low risk of privacy, low required network bandwidth, and low response time. The deployment scheme tackled insufficient support of deep learning from the JavaScript community, by aggregating NCNN and \ac{WASM}. Our \ac{WearMask} solution is a general framework where the detection algorithm can be replaced with other light-weight deep learning models. During the COVID-19 epidemic, our \ac{WearMask} solution can monitor and remind people to Wear masks, alleviating respiratory infections.



\ifCLASSOPTIONcaptionsoff
  \newpage
\fi



\bibliographystyle{IEEEtran}
\bibliography{main}
%







\end{document}